# INDUCTIVE LEARNING FOR RULE GENERATION FROM ONTOLOGY

Olegs Verhodubs

oleg.verhodub@inbox.lv

*Abstract*—**This paper presents an idea of inductive learning use for rule generation from ontologies. The main purpose of the paper is to evaluate the possibility of inductive learning use in rule generation from ontologies and to develop the way how this can be done. Generated rules are necessary to supplement or even to develop the Semantic Web Expert System (SWES) knowledge base. The SWES emerges as the result of evolution of expert system concept toward the Web, and the SWES is based on the Semantic Web technologies. Available publications show that the problem of rule generation from ontologies based on inductive learning is not investigated deeply enough.**

*Keywords- artificial intelligence; expert systems; semantic web; ontologie;, inductive learning.*

## I. INTRODUCTION

The final goal of the research is to develop a SWES (Semantic Web Expert System) that will be able to look for the ontologies in the Web according to a user request, to generate rules from the found ontologies, to supplement the SWES knowledge base with the generated rules and also to infer, based on a user request and the rules from the SWES knowledge base [1]. Ontologies are identified as the main resource for the SWES functioning [2], [3], however, to be precise, exactly rules instead of ontologies are the main resource for the SWES functioning. Ontology is raw material for the generation of rules, which is used in the SWES. In general, any source of information may serve as raw material for the generation of rules, and any source of information may be used as raw material for the generation of rules in the SWES, if it is formalized in one of the computer languages of information representation. The task of rule generation for the SWES from different sources of information, which are not ontologies, is one of the future tasks in the area of the SWES development.

This paper concerns the generation of rules from ontology. As opposed to the previous papers [2], [3], where the task of rule generation from ontology is solved by means of identifying the fragments of code and corresponding rules, generated from these fragments of code, this paper examines fundamentally new way of rule generation from ontology, using inductive learning. It is expected that the use of inductive learning will ensure deeper processing of raw materials for rules, where raw materials are ontologies. In addition, inductive learning based mechanism for rule generation is more technologically advanced, because it does not require the introduction of new fragments of code and corresponding rules in a rule generation routine. The main purpose of this paper is to evaluate the possibility of inductive learning use in rule generation from ontologies and to develop the way how this can be done. The paper is more theoretical, and it does not involve solving the task from the perspective of practice, because on the one hand it is a technical task, but on the other hand this is not the primary task for implementation in the SWES.

Mentioning ontologies, OWL (Web Ontology Language) ontologies are meant. The OWL language builds on RDF (Resource Description Framework), which is a data modeling language, and provides mechanisms for creating all the components of ontology: concepts, instances, properties (relations) and axioms [4]. Two sorts of properties can be defined in the OWL language. They are object properties and datatype properties. Object properties relate instances to instances. Datatype properties relate instances to datatype values, for example numbers or strings. Concepts are organized in hierarchy, having superconcepts and subconcepts. Hierarchy of concepts provides a mechanism for subsumption reasoning and inheritance of properties. In turn, axioms are used to provide information about classes and properties, for example to specify the equivalence of two classes or the range of a property [4].

This paper is structured as follows. The next section describes the way of inductive learning use in the task of rule generation from OWL ontologies. The third section presents conclusions.

## II. THE PROPOSED APPROACH

Mitchell defines machine learning as the field that is concerned with the question of how to construct computer programs that automatically improve with experience [5]. Machine learning can be defined as the ability of a computer program to improve its own performance, based on the past experience, by generation of a new data structure that is different from an old one, like production rules from input numerical or nominal data [6]. According to Ron Kovahi machine learning is the field of scientific study that concentrates on induction algorithms and on other algorithms that can be said to "learn" [7]. He also notes that machine learning is most commonly used to mean the application of induction algorithms, which is one step in the knowledge discovery process [7]. Induction infers generalized information, or knowledge, by searching for regularities among the data. Learning by induction

(inductive learning) is a search for a correct hypothesis/rule, or a set of them, which is guided by the given examples [6]. Induction algoritms, which are also called inductive learning algorithms, are algorithms that embody inductive learning.

Machine learning algorithms are domain independent [8], therefore inductive learning algorithms are domain independent, too. Since inductive learning algorithms are domain independent, it is possible to put forward the idea that inductive learning algorithms can be used for rule generation from OWL ontologies. It is necessary to represent OWL ontology in an appropriate form for the inductive learning algorithm in order to generate rules in this way. Ontology, which is represented by its categorized parts, is such an appropriate form for the inductive learning algorithm. Ontology parts can be categorized as follows:

- classes,
- properties,
- incoming relations,
- outgoing relations.

It is necessary to explain the difference between incoming and outgoing relations. Incoming are relations, which are going into the class. Outgoing are relations, which are going from the class. In turn, properties are the properties, which belong to the class. In this way the data set, obtained from ontology looks like as follows:

TABLE I. DATA SET FROM ONTOLOGY.

| Nr. | Class | Property | Incoming relation | Outgoing relation |
|---|---|---|---|---|
| 1 | $C_1$ | $P_1$ | $I_1$ | $O_1$ |
| 2 | $C_2$ | $P_2$ | $I_2$ | $O_2$ |
| 3 | $C_3$ | $P_3$ | $I_3$ | $O_3$ |
| … | … | … | … | … |
| n | $C_n$ | $P_n$ | $I_n$ | $O_n$ |

It is necessary to demonstrate some example in order to explain how the idea of inductive learning algorithm use for rule generation from OWL ontology can be realized. Let us assume that there is OWL ontology, and class "House" with two incoming relations ("liveIn", "builds") and two outgoing relations ("partOf", "equivalentOf"), which belongs to this ontology (Fig. 1.):

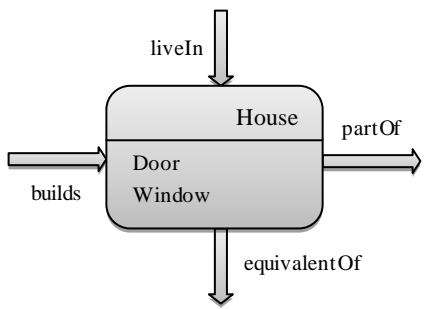

Fig. 1. Class with incoming and outgoing relations.

The data set, obtained from the class "House" with its properties ("Door", "Window") and relations ("builds", "liveIn", "partOf", "equivalentOf"), is the following:

TABLE II. DATA SET FROM CLASS "HOUSE".

| Nr. | Class | Property | Incoming relation | Outgoing relation |
|---|---|---|---|---|
| 1 | House | Door | liveIn | partOf |
| 2 | House | Door | builds | partOf |
| 3 | House | Door | liveIn | equivalentOf |
| 4 | House | Door | builds | equivalentOf |
| 5 | House | Window | liveIn | partOf |
| 6 | House | Window | builds | partOf |
| 7 | House | Window | liveIn | equivalentOf |
| 8 | House | Window | builds | equivalentOf |

The data set from TABLE II is obtained from one class with its properties and relations, but any ontology usually contains a lot of classes, properties and relations. Therefore such data set has to be supplemented by other data, obtained from other classes, properties and relations of the same ontology. After this is done, the obtained data set is used by one of the inductive learning algorithm. The result of functioning of such an algorithm is a set of rules, which can be used by the Semantic Web Expert System.

Selection of a particular inductive learning algorithm for rule generation from the data set, obtained from ontology, is a technical question. Of course, the selection of such algorithm implies a large field for experiments in accordance with many parameters as efficiency and others, but this is not the purpose of this paper. It is possible to notice that these experiments can be started with BEXA family algorithms (BEXA, FuzzyBEXA and FuzzyBEXA II). The selection of these algorithms is due to gained experience with these algorithms in Riga Technical University, where they are widely used and tested, and also the advantage that FuzzyBEXA and FuzzyBEXA II algorithms are fitted for fuzzy data. The fuzzy kind of some BEXA family algorithms for the task of the rule generation from ontology is necessary owing to the fuzzy nature of the information, including ontologies, in the Web that directly affects the operation of the SWES. For example, the final ontology for rule generation is being obtained in the SWES by means of several ontologies, which are found in the Web according to a user request, and which are merged into single ontology. The process of ontology merging involves membership function assignment for ontology elements [10]. After the process of ontology merging and membership function assignment for ontology elements is finished, the process of fuzzy rule generation is started. The data set, obtained from the ontology, where the ontology elements have their membership functions, differs from the data set, obtained from the ontology, whose elements have no membership functions. Let us look at an example. For instance, there is a class, which has its properties and relations, and these properties and relations have their own membership functions that are designated by µ (Fig.2.):

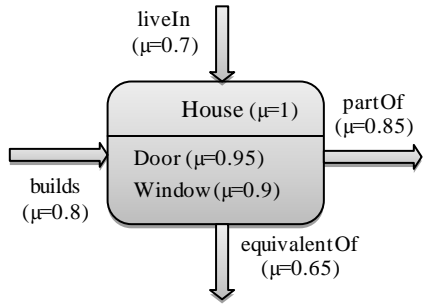

Fig. 2. Class and membership functions.

The data set that can be obtained from the class "House", properties "Door", "Window" and relations "builds", "liveIn", "partOf" and "equivalentOf" looks like as follows:

TABLE III. DATA SET WITH MEMBERSHIP FUNCTIONS.

| Nr. | Class | Property | Incoming relation | Outgoing relation |
|---|---|---|---|---|
| 1 | House (μ=1) | Door (μ=0.95) | liveIn (μ=0.7) | partOf (μ=0.85) |
| 2 | House (μ=1) | Door (μ=0.95) | builds (μ=0.8) | partOf (μ=0.85) |
| 3 | House (μ=1) | Door (μ=0.95) | liveIn (μ=0.7) | equivalentOf (μ=0.65) |
| 4 | House (μ=1) | Door (μ=0.95) | builds (μ=0.8) | equivalentOf (μ=0.65) |
| 5 | House (μ=1) | Window (μ=0.9) | liveIn (μ=0.7) | partOf (μ=0.85) |
| 6 | House (μ=1) | Window (μ=0.9) | builds (μ=0.8) | partOf (μ=0.85) |
| 7 | House (μ=1) | Window (μ=0.9) | liveIn (μ=0.7) | equivalentOf (μ=0.65) |
| 8 | House (μ=1) | Window (μ=0.9) | builds (μ=0.8) | equivalentOf (μ=0.65) |

The data set from TABLE III has to be supplemented with other data, obtained from other classes of the same ontology with their properties, relations and their membership functions, if these classes, properties and relations are in this ontology. In this regard the principle of the data set supplementation is the same as in the data set without membership functions. After the data set with membership functions is formed, one of inductive learning algorithms for fuzzy data as FuzzyBEXA, FuzzyBEXA II or others should be applied in order to get fuzzy rules of this type: IF x (μ = a) THEN y (μ = b), where x, y – statements, μ - membership function, a, b – values of membership functions. Generated fuzzy rules are stored in the SWES knowledge base and are used by the SWES semantic reasoner in order to produce new facts in accordance with the user request.

## III. CONCLUSION

This paper develops the idea of rule generation from OWL ontologies. If previous achievements in the area of rule generation from OWL ontologies are based on the identification of OWL code fragments, which may serve as sources for generation of a certain type of rules [2], [3], this research is based on a completely new principle, which consists in the use of inductive learning algorithms for rule generation from OWL ontology. The paper presents the evaluation of the inductive learning use possibility in rule generation from ontologies and the way how this can be done. The Bexa family algorithms are mentioned as possible inductive learning algorithms for rule generation from OWL ontologies.

It is necessary to notice that the use of inductive learning algorithms in the task of rule generation from OWL ontologies has its own advantages and disadvantages. The main advantage of this approach is that it allows us to generate all possible rules from one source of information, i.e. OWL ontology. In contrast to the previous approach, where it is necessary to set new schemes for rule generation from OWL ontology, the proposed approach requires the proper data set, and rules are generated in the process of inductive learning algorithm execution. In addition, if one inductive learning algorithm generates insufficient amount of rules, it is possible to apply different inductive learning algorithms to the same OWL ontology. Utilizing certain amount of such algorithms, it is possible to obtain all possible rules from OWL ontology. Despite the significant advantage, the approach has serious disadvantages. The first disadvantage is efficiency. For example, the OWL class (Fig. 2), which has two properties and four relations, forms the data set, which has eight tuple of data (Table III). Real OWL ontology from the Web may have hundreds or even thousands of classes, properties and relations. The use of only one inductive learning algorithm for one ontology may slow down the functioning of the SWES. This is the obstacle, which does not allow scheduling the task of rule generation "on the fly". The use of several inductive learning algorithms for the same ontology makes it an insurmountable obstacle. However the main disadvantage of inductive learning use for rule generation from OWL ontology is that this approach is too mechanistic and rigid. This means that rules are generated apart from the necessity. The use of inductive learning approach demands the full execution of the inductive learning algorithm to get needed rule. There is a need of more intelligent and selective approach, which would give an opportunity to manage the process of rule generation and to obtain exactly needed rule at particular moment. The approach of OWL code fragments and their corresponding rules is fully consistent with this parameter. Thus, this approach is preferable for the SWES, and therefore it will be implemented first of all.